\documentclass[10pt,twocolumn,letterpaper]{article}

\usepackage{wacv}
\usepackage{times}
\usepackage{epsfig}
\usepackage{graphicx}
\usepackage{amsmath}
\usepackage{amssymb}
\usepackage{url}

\wacvfinalcopy 

\def\wacvPaperID{****} 

\ifwacvfinal\fi
\begin{document}

\title{Deep Bayesian Network for Visual Question Generation}

  \author{ {Badri N. Patro}  \quad {Vinod K. Kurmi}  \quad {Sandeep Kumar} \quad  {Vinay P. Namboodiri} \\
  Indian Institute of Technology, Kanpur \\
  {\tt \{badri,vinodkk,sandepkr,vinaypn\}@iitk.ac.in} \\
}
  

\maketitle
\ifwacvfinal\thispagestyle{empty}\fi

\begin{abstract}
Generating natural questions from an image is a semantic task that requires using vision and language modalities to learn multimodal representations. Images can have multiple visual and language cues such as places, captions, and tags. In this paper, we propose a principled deep Bayesian learning framework that combines these cues to produce natural questions. We observe that with the addition of more cues and by minimizing uncertainty in the among cues, the Bayesian network becomes more confident. We propose a Minimizing Uncertainty of Mixture of Cues (MUMC), that minimizes uncertainty present in a mixture of  cues experts for generating  probabilistic questions. This is a Bayesian framework and the results show a remarkable similarity to natural questions as validated by a human study. We observe that with the addition of more cues and by minimizing uncertainty among the cues, the Bayesian framework becomes more confident. Ablation studies of our model indicate that a subset of cues is inferior at this task and hence the principled fusion of cues is preferred. Further, we observe that the proposed approach substantially improves over state-of-the-art benchmarks on the quantitative metrics (BLEU-n, METEOR, ROUGE, and CIDEr). Here we provide project link for Deep Bayesian VQG \url{https://delta-lab-iitk.github.io/BVQG/}.
\end{abstract}
\vspace{-0.8em}
\section{Introduction}

\begin{figure}[ht]
\centering
\includegraphics[width=0.91\columnwidth]{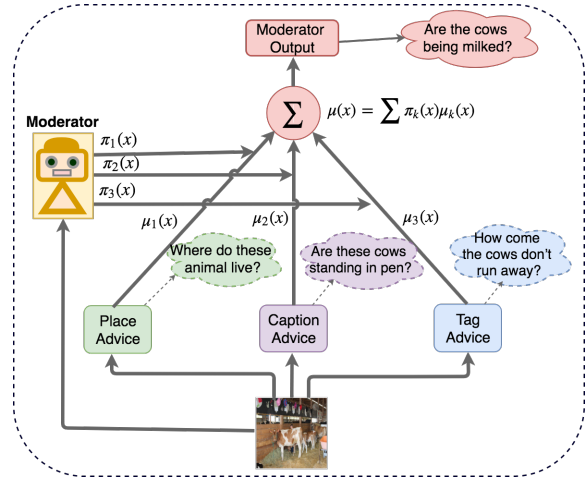}
\caption{Here we give an overview of our network. We have three experts which provide us with information (advice) related to different cues. These are shown as Place Expert, Caption Expert and Tag Expert respectively. Then we have a moderator which weighs these advices and passes the resultant embedding to the decoder to generate natural question.}
\label{fig:univerise_intro}
\vspace{-2em}
\end{figure}
The interaction of humans and automated systems is an essential and increasingly active area of research. One such aspect is based on vision and language-based interaction. This area has seen many works related to visual question answering~\cite{VQA} and visual dialog~\cite{visdial}. 
 Current dialog systems as evaluated in~\cite{visdial_eval} show that when trained between bots, AI-AI dialog systems show improved performance, but that does not translate to actual improvement for Human-AI dialog. This is because, the questions generated by bots are not natural and therefore do not translate to improved human dialog. Therefore it is imperative that improvement in the quality of questions will enable dialog agents to perform well in human interactions. Further, in~\cite{GanjuCVPR17} the authors show that unanswered questions can be used for improving VQA, Image captioning and Object Classification. So the generation of natural questions will further improve performance on these tasks.
While not as well studied as the other tasks of answering questions or carrying a conversation, there has been work aimed at generating natural and engaging questions from an image~\cite{mostafazadeh2016generating, jain2017creativity} which is the VQG task. The underlying principle for all these methods is an encoder-decoder formulation. We argue that there are underlying cues that motivate a natural question about an image. It is essential to incorporate these cues while generating questions. For each image, there may be a different underlying cue that is most pertinent. For some images, the place may be important (`Is it a cowshed?') whereas for others the subject and verb may provide more context (`Are the horses running?'). Our work solves this problem by using a principled approach for multimodal fusion by using a {\em mixture of experts (MoE)} model to combine these cues. We hypothesize that the joint distribution posterior based on the cues correlates with natural semantic questions.

To verify our hypothesis, we systematically consider approaches to extract and combine descriptors from an image and its caption.
We argue that some of the critical descriptors that could provide useful context are: a) Location description, b) Subject and Verb level description and c) Caption level description.
\vspace{-0.7em}
\begin{itemize}
    \item {\em Location description}: For certain kinds of images that involve locations such as train-stations or bus-stations, the context is dominated by location. For instance, natural questions may relate to a bus or a train and hence could be more related to the destination or time related information. In such scenarios, other cues may be secondary cues. In our work, we obtain a posterior probability distribution that captures the probability of the location cue by training a Bayesian deep CNN.
    \item {\em Subject and Verb level description}: In certain images, the main context may relate to the subject and verb (for instance, food and eating). In such cases, subject-verb combinations dominate the context. Given an image we obtain a posterior probability distribution over the set of tags. 
     \item {\em Caption}: For a set of natural questions, an important context could be obtained from an image caption. We can now use state-of-the-art image captioners to generate descriptive captions of an image, which is useful information for generating questions pertinent to the same image. We use this information by obtaining a posterior distribution on the caption generator. 
\end{itemize}

\begin{figure}[ht]
\centering
\vspace{-0.7em}
\includegraphics[width=\columnwidth]{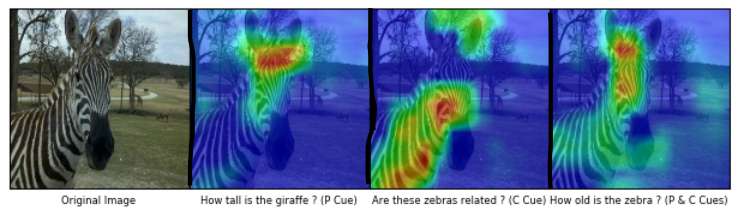} 
\caption{Here we visualize the GradCAM maps corresponding to single and multiple cues for question generation.
} 
\vspace{-0.7em}
\label{fig:motivation}
\end{figure}

We show the GradCAM~\cite{selvaraju2017grad} visualisations for the questions generated on the basis of single and multiple cues in Figure~\ref{fig:motivation}. We see that the model focuses on different regions when provided single cues (Place and Caption in the second and third image in Figure~\ref{fig:motivation}) and asks poor questions, but when we provide both the Place and Caption cues to the model, it focuses on correct regions which results in sensible question. So incorporating multiple cues through a principled approach in our model should lead to more natural questions. 

We combine these distributions (cues) to estimate latent distributions which are then mixed through a moderator network and used by a decoder module to generate questions.
On obtaining these distributions, we then obtain the combination of the cues that provides us with a combined latent distribution that is used by a decoder module that generates the question.  The approach is illustrated in figure~\ref{fig:univerise_intro}. The main aspect that we focus on this paper is to investigate a number of cues that can provide us with the necessary semantic correlation that can guide generation of natural questions and the ways in which these cues can be combined.
The contributions of this paper are as follows: 
 \begin{itemize}
  \item We provide Bayesian methods for obtaining posterior distributions by considering the advice of various experts that capture different cues embedding and aid in generating more natural questions.
  \item We propose a method to capturing and minimizing uncertainty (aleatoric and epistemic) in question generation task. 
\item We show that by Minimizing Uncertainty in Multiple Cues (MUMC) method with the help of Gaussian cross-entropy and variance minimizing loss, improves the score.
\item  We also analyze the different ablations of our model and show that while each of these cues does affect the generation, a probabilistic combination of these improves the generation in a statistically significant way.

 \end{itemize}

\begin{figure*}[ht!]
\centering
\includegraphics[width=0.9\textwidth]{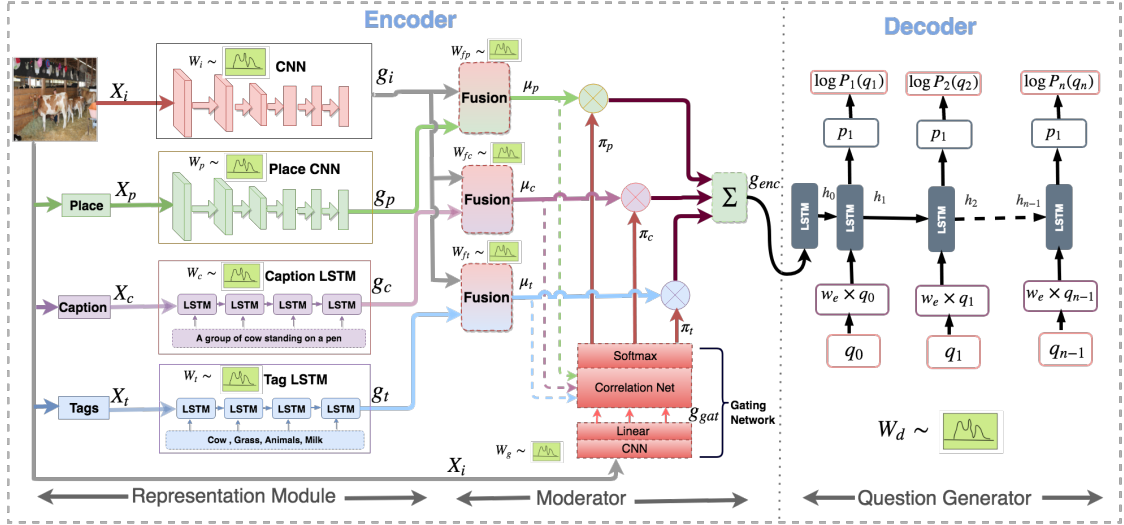}
\caption{Multi-Cue Bayesian Moderator Network. We first use a Bayesian CNN/LSTM to obtain the embeddings $g_i,g_p,g_c,g_t$ and then fuse those using the Fusion Module to get $\mu_p,\mu_c,\mu_t$.
These embeddings are then passed to the Moderator network. These are then fed to the decoder to get the questions for each image. }
\label{fig:univerise_model}
\vspace{-1em}
\end{figure*}


\section{Related Work}
\label{sec:lit_surv}
The task of automatically generating questions is well studied in the NLP community, but it has been relatively less explored for generating image related questions. 
On the other hand, there has been extensive work done in the Vision and Language domain for solving image captioning~\cite{Barnard_JMLR2003,Farhadi_ECCV2010,Kulkarni_CVPR2011,Socher_TACL2014,Vinyals_CVPR2015,Karpathy_CVPR2015,Xu_ICML2015,Fang_CVPR2015,Chen_CVPR2015,Johnson_CVPR2016,Yan_ECCV2016}, Visual Question Answering (VQA)~\cite{Malinowski_NIPS2014,Lin_ECCV2014,VQA,Ren_NIPS2015,Ma_AAAI2016,Noh_CVPR2016,Fukui_arXiv2016,Yang_CVPR2016,Kim_ICLR2017,Patro_2019_ICCV} and Visual Dialog~\cite{visdial,Aytar_TPAMI2017,Velivckovic_SSCI2016,Vijayakumar_2016diverse,Yu_ICCV2015}.
However, Visual Question Generation (VQG) is the task aimed at generating `natural and engaging' questions for an image and was proposed by Mostafazadeh \textit{et al.}~\cite{mostafazadeh2016generating}. It focuses more on questions which are interesting for a person to answer and not on those which can be answered simply by looking at the image and hence could be used to evaluate a computer vision model. 
One of the works in this area is~\cite{Yang_arXiv2015} where the authors proposed a method for continuously generating questions from an image and subsequently answering the questions being generated.
In~\cite{mostafazadeh2016generating}, the authors used an encoder-decoder based framework that has been further adopted in our work by considering various contexts.
In~\cite{jain2017creativity}, the authors extend it by using a Variational Autoencoder based sequential routine to obtain natural questions by performing sampling of the latent variable. In a very recent work by~\cite{patro2018multimodal}, the authors use an exemplar based multimodal encoder-decoder approach to generate natural questions.
Our work extends our previous work ~\cite{patro2018multimodal} by proposing a deep Bayesian multimodal network that can generate multiple questions for an image. 

It has been shown that for small datasets, 
Bayesian Neural Networks~\cite{Gal_ARX2015} are robust to overfitting and weights are easily learned.
 The earliest works in Bayesian Neural networks by~\cite{neal1993bayesian,neal2012bayesian,mackay1992bayesian,denker1987large,denker1991transforming,tishby1989consistent,buntine1991bayesian} focused on the idea that model weights come from a random distribution and tried to approximate the posterior distribution of the weights given the data. 
To approximate the intractable posterior distribution, variational inference is one of the existing approaches introduced by~\cite{Hinton_ACM1993,Barber_NATO1998,Graves_NIPS2011,Blundell_ARX2015}.
Gaussian distribution is a popular choice for the variational distribution, but it is computationally expensive~\cite{Blundell_ARX2015}. This can be overcome by using a Bernoulli distribution which we also use in our work. 
 There has been some recent work which applies these concepts to CNNs~\cite{Gal_ARX2015} (Bayesian CNN) and LSTMs~\cite{Gal_NIPS2016} (Bayesian LSTM) for obtaining probabilistic representations of images and sequential data respectively. These methods show that using Dropout~\cite{srivastava2014dropout} training in deep neural networks (DNN) can be interpreted as an approximate Bayesian inference in deep Gaussian processes and can be used to represent uncertainty in DNNs. Recently Kurmi {\it et al.}~\cite{kurmi_cvpr2019attending} has proposed a method to minimise uncertainty in source and target domain and Patro {\it et al.}~\cite{Patro_2019_ICCV} has proposed an gradient based method to minimise uncertainty in the attention regions for solving VQA task.
 To the best of our knowledge, the usage of Bayesian fusion of cues for end-to-end inference setting has not been considered previously for a deep learning setting. Having a principled approach for fusing multiple cues will be beneficial even in other settings such as autonomous robots, cars, etc.
We compare our work with the some related works for question generation in the experimental section and show that considering different contexts and combining them using a product of experts setup can improve the task of natural question generation.

\section{Method}
We adopt a generation framework that uses an image embedding combined with various cues namely, place, caption and tag embeddings to generate natural questions. We propose a Multi Cue Bayesian Moderator Network (MC-BMN) to generate questions based on a given image. 
\subsection{Finding Cues}
\vspace{-0.2cm}
As location is one of an important cue, we used different scene semantic categories present in the image as a place-based cue to generate natural questions. We use pre-trained PlaceCNN~\cite{Zhou_PAMI2017} which is modeled to classify 365 types of scene categories. Captions also play a significant role in  providing semantic meaning for the questions for an image. 
Tags provide information relevant to various topics in an image.   
We are using parts-of-speech (POS) tagging for captions to obtain these. The tags are clustered into three categories namely, Noun tag, Verb tag and Question tags. Noun tag consists of all the noun \& pronouns present in the caption, and similarly, the Verb tag includes verb \& adverbs present in the caption sentence whereas the Question tags consist of (Why, How, What, When, Where, Who and Which). 
Each tag token is represented as a one-hot vector of the dimension of vocabulary size. For generalization, we have considered five tokens from each category of the tags.
\subsection{Representation module}
\vspace{-0.2cm}
Given an input image $x_i$, we obtain its embedding $g_i$ using a Bayesian CNN~\cite{Gal_ARX2015} that we parameterize through a function $G(x_i,W_i)$ where $W_i$ are the weights of the Bayesian CNN. We have used a pretrained VGG-19~\cite{simonyan_arxiv2014} CNN trained on ImageNet for image classification task as the base CNN which was also used by the previous state-of-the-art methods like~\cite{mostafazadeh2016generating} and~\cite{jain2017creativity}. To make Bayesian CNN~\cite{Gal_ARX2015}, We use pretrained CNN layers and put Dropout layer with dropout rate $p$, before each CNN layer to capture Epistemic Uncertainty. Then, we extracted $g_i$, a d-dimensional image feature from the Bayesian CNN network as shown in figure~\ref{fig:univerise_model}.
Similarly we obtain place embeddings $g_p$ using a Bayesian PlaceCNN $G(x_p,W_p)$ for place input $x_p$. The Bayesian PlaceCNN is the pretrained PlaceCNN with similar placement of dropout layer as the VGG-19 CNN.\\
To generate caption and tag embeddings, we use a $V$ (size of vocabulary) dimensional one-hot vector representation for every word in the Caption \& Tags and transform them into a real valued word embedding $X_{we}$ for each word using a matrix $W_C \in \mathcal{R}^{E_{C}\times V}$.
Then the $E_C$ dimensional word embeddings are fed to the Bayesian LSTM to obtain the required representations for the caption and tag inputs. Bayesian LSTM is designed by adding dropout layer into each gate of the LSTM and output layer of the LSTM as done in~\cite{Gal_NIPS2016}. So we obtain $g_c,g_t$ using a Bayesian LSTMs $ F(x_c,W_c)$ and $F(x_t,W_t)$ for caption input $x_c$, and tag input $x_t$ respectively.




\subsection{Bayesian Fusion Module }\label{Joint_Cues}
\vspace{-0.2cm}
There have been some works for VQA which use a projection of multiple modalities to a common space with the help of a fusion network to obtain better results~\cite{VQA,Zhou_arXiv2015}.
We use a similar fusion network to combine multiple modalities, namely caption, tag and place with the image.
The fusion network can be represented by the following equations: 
\[\mu_p =  W_{pp} \ast\tanh( W_{i} g_{i}  \otimes W_{p}  g_{p}  + b_p)\]
\[\mu_c =  W_{cc} \ast\tanh( W_{i} g_{i}  \otimes W_{c}   g_{c}  + b_c)\]
\[\mu_t =  W_{cc} \ast\tanh( W_{i} g_{i}  \otimes W_{t}   g_{t}  + b_t)\]
where, $g_\star$ is the embedding for corresponding cues, $W_{\star}$ and $b_{\star}$ are the weights and the biases for different cues(${\star} \text{ represent} \{p,c,t\} $). Here $\otimes$ represent element-wise multiplication operation. We use a dropout layer before the last linear layer for the fusion network. We also experimented with other fusion techniques like addition, attention, and concatenation but element-wise multiplication performed the best for all the metrics. 

\subsection{Bayesian Moderator Module}
\vspace{-0.2cm}
We propose a Moderator Module to combine the fused embeddings. The proposed model is similar to the work of~\cite{Ueda_IEEE2000,Baldacchino_MSSP2016,Yuksel_IEEETransNNLS2012}. The Moderator module receives input image $x_i$ and obtains a gating embedding $g_{gat}$ using a Bayesian CNN that we parametrize through a function $G(x_i,W_{g})$. Then, a correlation network finds the correlation between gating embedding $g_{gat}$ and $\mu_{B}$ to obtain scaling factors $\pi_B$, where $B \in \{p,c,t\}$. Finally, Moderator combines the fused embeddings $\mu_{B}$ with the scaling factors $\pi_B$ to obtain the final embedding $g_{enc}$.
\[ g_{gat}=BayesianCNN{(x_i;W_g)}\]
\[ \pi_B= softmax(g_B * g_{gat}) \forall B \in \{p,c,t\}\]
\[ g_{enc}=\sum_{B \in \{p,c,t\}}{\pi_B*\mu_B} \]


 \begin{figure*}[htb]
     \small
     \centering
     \begin{tabular}[b]{ c  c  c}
     (a) MC-BMN & (b) MC-BMN-2 & (c) MC-SMix \\ 
     \includegraphics[width=0.33\textwidth]{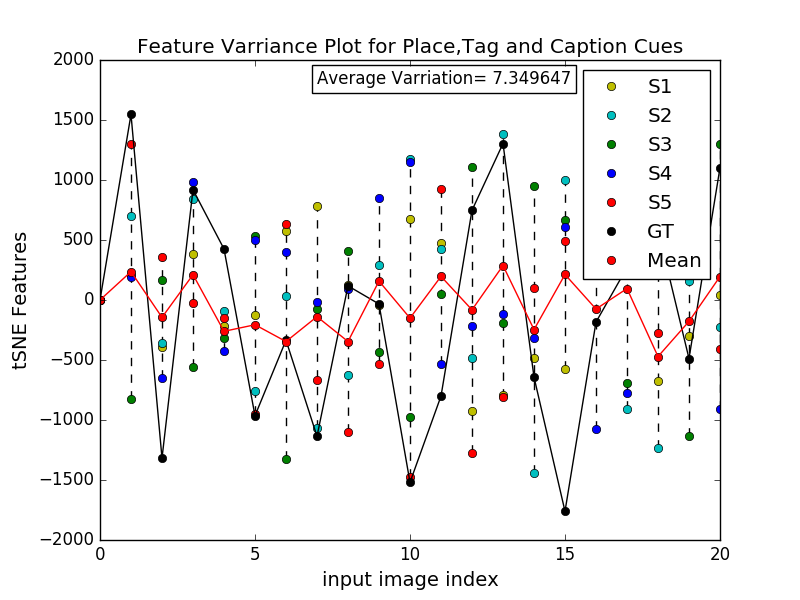}
     & \includegraphics[width=0.33\textwidth]{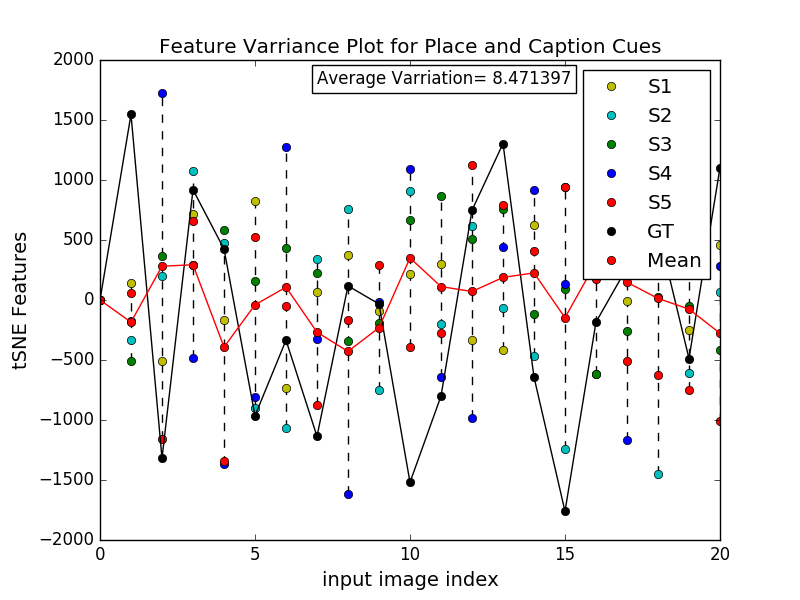}
     & \includegraphics[width=0.33\textwidth]{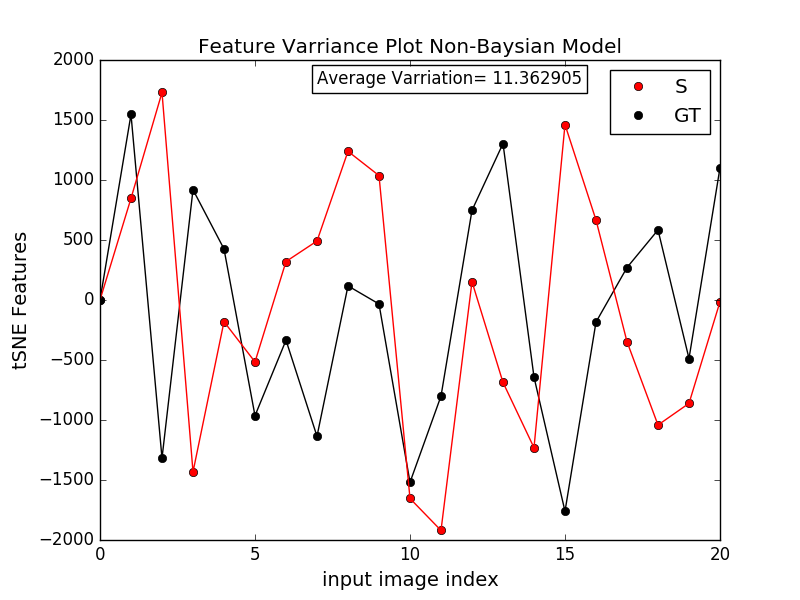}
       \end{tabular}
      \caption{Variance plots for Bayesian and Non-Bayesian networks for a toy example of 20 images. We have drawn 5 samples of each image using Monte-Carlo sampling from a distribution (this is predictive posterior distribution for the Bayesian case) and then plot the mean features of these 5 samples along with the ground truth features. MC-BMN (3 cues) reduces normalized variance (difference in mean feature value \& ground truth feature value) as compared to two cues(MC-BMN-2). Whereas for MC-SMix(Non-Bayesian network), the variance is too high as compared to MC-BMN.}
      \label{tbl:variance}
      \vspace{-1.8em}
 \end{figure*}

\subsection{Decoder: Question Generator}
\vspace{-0.2cm}


The decoder's task is to predict the whole question sentence given an image $I$ and its cues (C). The probability for a question word depends on the previously generated words. This conditional probability $P( q_{t+1} | {I ,C},q_0,...,q_{t})$ is modeled with a LSTM for sequential tasks such as machine translation~\cite{Sutskever_NIPS2014}. We use a Bayesian LSTM similar to the one used in our Representation Module for this question generation task.
At $t$ =$-1$, we feed the moderator advice $g_{enc}$ to the LSTM. The output of the word with maximum probability in the distribution $P( q_t | {g_{enc}},h_{t})$ in the LSTM cell at time step $t$ is fed as input to the LSTM cell at step $t$+$1$ as mentioned in the decoder in figure~\ref{fig:univerise_model}. At time steps $t=0:(T-1)$, the softmax probability is given by:
\begin{equation}
 \begin{split}
& x_{-1}=g_{enc} \\
& x_t=W_C*q_t,  \forall t\in \{0,1,2,...T-1\} \\
& {h_{t+1}}=LSTM(x_t,h_{t}), \forall t\in \{0,1,2,...N-1\}\\
& o_{t+1} = W_o * h_{t+1} \\
& \hat{y}_{t+1} = P( q_{t+1} | g_{enc},h_{t})= \mbox{softmax}(o_{t+1})\\
& Loss_{t+1}=\mbox{loss}(\hat{y}_{t+1},y_{t+1})
 \end{split}
\end{equation}
where $h_t$ is the hidden state and $o_t$ is the output state for LSTM.
\vspace{-0.1cm}
\subsubsection{Uncertainty in Generator Module}
\vspace{-0.3cm}
The decoder module is generating diverse words which lead to uncertainty in the generated sentences. The uncertainty  present in the model can be captured by estimating Epistemic uncertainty~\cite{kendall2015bayesian}, and the uncertainty  present in the data can be captured by estimating Aleatoric uncertainty~\cite{Gal_ICML2016}. The predictive uncertainty~\cite{malinin2018predictive} is the total uncertainty which is the combination of both uncertainties. The predictive uncertainty measures the model's capability for generating question word token by focusing on various cues (caption, tag, and place) networks.  We use the similar Bayesian decoder network to capture predictive uncertainty by approximating the posterior over the weights of Bayesian decoder using MC-dropout as described in~\cite{Kendall_CVPR2018,kurmi_cvpr2019attending,Patro_2019_ICCV}. The uncertainty in these cues moderators occurs mainly due to either noise or lack of data to learn mixture of cues. We proposed a method Minimising Uncertainty for mixture of Cue (MUMC), which enhances model performance by minimizing uncertainty.
\vspace{-0.4em}
\begin{figure}[ht]
\centering
\includegraphics[width=1.05\columnwidth]{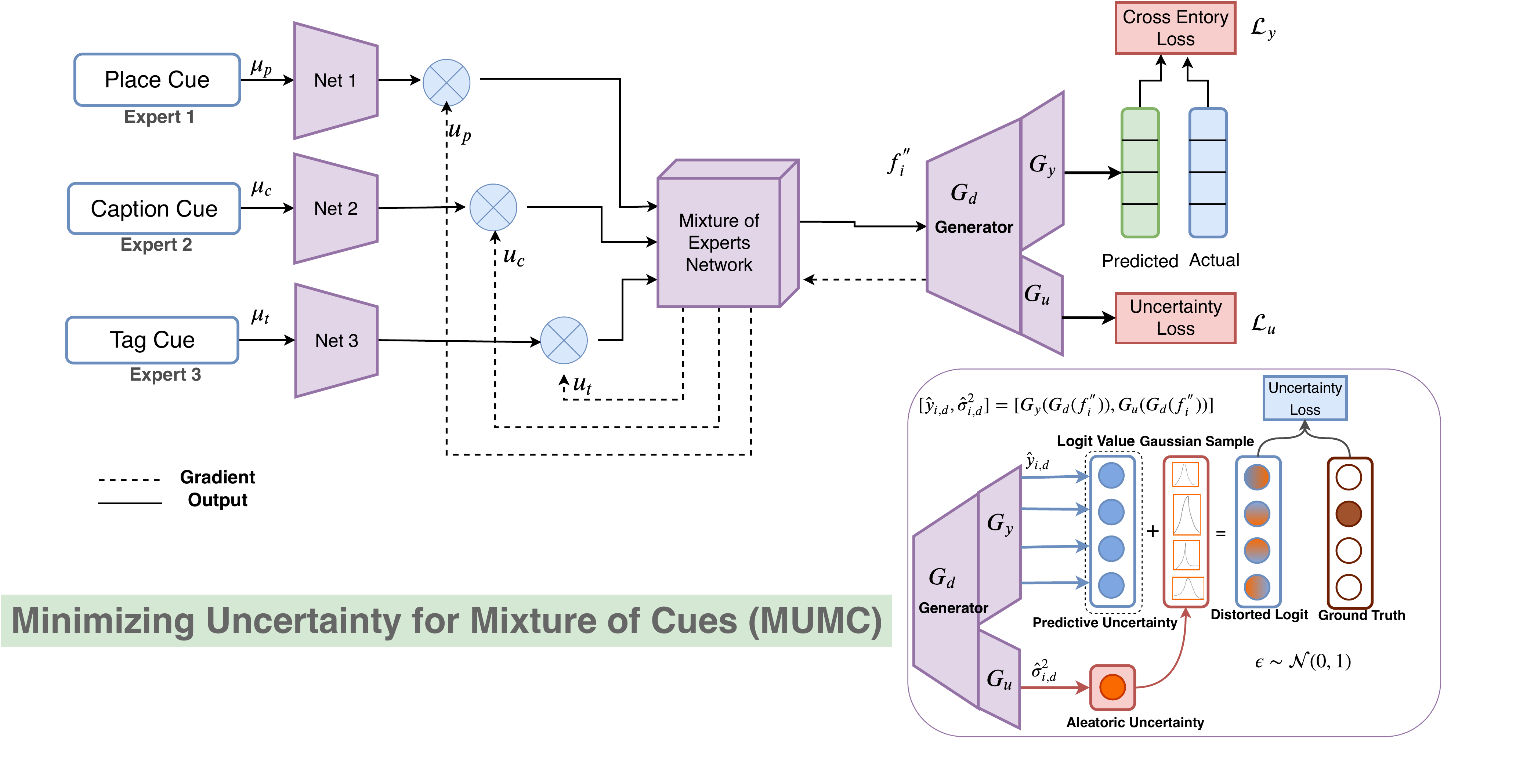}
\vspace{-0.5em}
\caption{Model architecture for minimizing uncertainty for mixture of Cues}
\label{fig:variance_net}
\vspace{-1.2em}
\end{figure}

 \begin{figure*}[htb]
     \small
     \centering
     \begin{tabular}[b]{ c  c  c}
     (a) COCO & (b) BING & (c) FLICKR \\ 
     \includegraphics[width=0.33\textwidth]{coco_1.pdf}
     & \includegraphics[width=0.33\textwidth]{bing_1.pdf}
     & \includegraphics[width=0.33\textwidth]{flickr_1.pdf}
       \end{tabular}
      \caption{Sunburst plot of generated questions for MC-BMN  on VQG-COCO dataset,VQG-Bing dataset, VQG-Flickr dataset  are shown in Fig-a, Fig-b, Fig-c respectably : The $i^{th}$ ring captures the frequency distribution over words for the $i^{th}$ word of the generated question. While some words have high frequency, the outer rings illustrate a fine blend of words.}
      \label{fig:sunburst}
       \vspace{-1.05em}
 \end{figure*}

\textbf{MUMC:}
The decoder generates a logit out $y_{i,g}$ and variance network predict variance in each generated word token.
\begin{equation}
\label{e5}
y_{i,g}= G_y(G_o(f_i)), \quad v_{i,g}= G_v(G_o(f_i)) 
\end{equation}
\text{where} $f_i = g_{gen}$ is the output feature of the Bayesian Moderator Module. $G_o$ is the decoder network, $G_y$ is the final word token classifier and $G_v$ is the variance predictor network. In order to capture uncertainty in the data, we learn observational noise parameter $\sigma_{i,g}$ for each input point $x_i$ and its cues. This can be achieved by corrupting the logit value ($y_{i,g}$) with the Gaussian noise with variance $\sigma_{i,g}$ (diagonal matrix with one element for each logits value) before the softmax layer. We defined a Logits Reparameterization Trick (LRT), which combines two outputs $y_{i,g},\sigma_{i,g}$ and then we obtain a loss with respect to the ground truth. That is, after combining we get $\mathcal{N}(y_{i,g},(\sigma_{i,g})^2)$ which is expressed as:
\begin{equation}
\label{e4}
\hat{y}_{i,t,g}=y_{i,g}+\epsilon_{t,g}\odot  \sigma_{i,g},\quad where \quad \epsilon_{t,g} \sim \mathcal{N}(0,1)
\end{equation}
\begin{equation}
    \label{e6}
  \mathcal{L}_u= \sum_{i}\log \frac{1}{T}\sum_{t}{\exp{(\hat{y}_{i,t,g} - \log \sum_{M^`}\exp{\hat{y}_{i,t,{M^`}}})}}
\end{equation}
Where M is the total word tokens, $\mathcal{L}_u$ is minimized for true word token $M$, and T is the number  of  Monte  Carlo  simulations. $M^{'}$ is the element in the logit vector $y_{i,t}$  for all the classes.
$\sigma_{i,g}$ is the standard deviation, ( $\sigma_{i,g}=\sqrt{v_{i,g}}$). 

We compute gradients of the predictive uncertainty $\sigma_g^2$ of our generator with respect to the features $f_i$. We first compute gradient of the uncertainty loss  $\mathcal{L}^v$ with respect to cues moderator feature $f_i=g_{gen}$ i.e. $\frac{\partial{\mathcal{L}_v}}{\partial{f_i}}$  Now we pass the uncertainty gradient through a gradient reversal layer to reverse the gradient of the all the cues is given by \[\nabla_{y} = -\gamma * \frac{\partial{\mathcal{L}_u}}{\partial{f_i}}\]
We perform a weighted combination of forward cues moderator feature maps ${\mu_p,\mu_c,\mu_t}$ with the reverse uncertainty gradients i.e.
\[ \nabla^{'}_{g_{enc}}=\sum_{B \in \{p,c,t\}}{ -\gamma * \frac{\partial{\mathcal{L}_u}}{\partial{f_i}}*\mu_B} \]
We use residual connection to obtain the final moderator cue feature by combining original cue moderator feature with  the gradient certainty mask $\nabla_{y}^{''}$ and is given by:
\[ g^{'}_{enc}= g_{enc} + \sum_{B \in \{p,c,t\}}{\nabla^{'}_{g_{enc}} *g_{enc}} \]
From this moderator feature we are generating question word tokens.
\subsection{Cost Function} \label{costfun_ale}
\vspace{-0.5em}
We estimate aleatoric uncertainty in logit space by distorting each logit value by the variance obtained from data. The uncertainty present in each logit value can be minimized using cross-entropy loss on Gaussian distorted logits as shown in equation-~\ref{e4}. The distorted logits is obtained using Gaussian multivariate function with positive diagonal variance. The uncertainty distorted loss is the difference between actual cross entropy loss and the uncertainty loss mentioned in equation-~\ref{e6}. The difference is passed through an activation function to enhance the difference in either direction and is given by :
 \begin{equation}
  \mathcal{L}_{u}=\begin{cases}
    \alpha (\exp^{[\mathcal{L}_{p}-\mathcal{L}_{y}]}-1), & \text{if $[\mathcal{L}_{p}-\mathcal{L}_{y}]<0$}.\\
    [\mathcal{L}_{p}-\mathcal{L}_{y}], & \text{otherwise}.
  \end{cases}
\end{equation}
The final cost function for the network combines the loss obtained through uncertainty (aleatoric or predictive) loss $\mathcal{L}_v$ for the attention network with the cross-entropy.

In the question generator module, we use the cross entropy loss function between the predicted and ground truth question, which is given by:
 \begin{equation}
L_{gen}=\frac{-1}{NM}\sum_{i=1}^{N} \sum_{t=1}^{M} {y_{t} \log {p}(q_{t}|(g_{enc})_i,{q_0,..q_{t-1}})}
\end{equation}
where, $N$ is the total number of training examples, $M$ is the total number of question tokens, $\text{P}(q_{t}|(g_{enc})_i,{q_0,..q_{t}})$ is the predicted probability of the question token, $y_t$ is the ground truth label.
We have provided the pseudo-code for our method in our project webpage.

\vspace{-0.2cm}
\section{Experiments}{\label{experiments}}
\vspace{-0.2cm}
We evaluate the proposed method in the following ways: First, we evaluate our proposed MC-BMN against other variants described in section~\ref{variants}. Second, we further compare our network with state-of-the-art methods such as Natural~\cite{mostafazadeh2016generating} and Creative~\cite{jain2017creativity}. Third, we have shown in figure~\ref{tbl:variance}, the variance plots for different samples drawn from the posterior for Bayesian  and Non-Bayesian methods. Finally, we perform a user study to gauge human opinion on the naturalness of the generated question and analyze the word statistics with the help of a Sunburst plot as shown in Figure~\ref{fig:sunburst}. We also consider the significance of the various methods for combining the cues as well as for the state-of-the-art models. The quantitative evaluations are performed using standard metrics namely BLEU~\cite{Papineni_ACL2002}, METEOR~\cite{Banerjee_ACL2005}, ROUGE~\cite{Lin_ACL2004} and CIDEr~\cite{Vedantam_CVPR2015}. BLEU metric scores show strong correlations with human for the VQG task and is recommended by Mostafazadeh \textit{et al.}~\cite{mostafazadeh2016generating} for further bench-marking.
In the  paper, we provide the comparison with respect to only BLEU-1 and METEOR metrics and the full comparison with all metrics(BLEU-n, CIDER and ROUGE) and further details are present in our project webpage\footnote{\url{https://delta-lab-iitk.github.io/BVQG/}\label{webpage}}. 

\begin{table}[ht]
\small
\centering
\begin{tabular}{|l|lccc|}
\hline \bf Method &  \bf BLEU1  &\bf METEOR & \bf ROUGE & \bf CIDEr \\ \hline 
MC-SMix & 31.1   &19.1  &32.6 & 42.8\\
MC-BMix  & 36.4  &  22.6 & 40.7 & {46.6}\\
MC-SMN & 33.1   &21.1  &37.6 & 47.8\\
MC-BMN +PC & {24.6}   &{11.1}& {24.0} & 45.2\\
MC-BMN & \bf{40.7}  &  \textbf{22.6} & \textbf{41.9} & \textbf{49.7}\\\hline
\end{tabular}
\caption{\label{full_score_tab_1}Ablation Analysis on VQG-COCO Dataset.It has the different variations of our model described in `Comparison with State-of-the-Art and Ablation Analysis' section of the paper. As expected the performance with the generated captions is not as good as with the ground truth captions. Note that these are the max scores over all the epochs. PC tends for Predicted Caption
}
\end{table}


\begin{table*}
  \centering
  \begin{tabular}{|l|cc|cc|}
    \hline
    {\textbf{Methods}} & \multicolumn{2}{c|}{\textbf{BLEU1}} &   \multicolumn{2}{c|}{\textbf{METEOR}}\\
    \cline{2-5}
    & \textbf{Max} & \textbf{Avg} & \textbf{Max} & \textbf{Avg}\\
    \hline
        Natural~\cite{mostafazadeh2016generating}  & 19.2 & - & 19.7& -  \\
        Creative~\cite{jain2017creativity}        & 35.6    & -  & 19.9 & -\\
        MDN~\cite{patro2018multimodal} &36.0&-&\textbf{23.4}&-\\
        \hline
        Img Only (Bernoulli Dropout (BD))         & 21.8 & 19.57 $\pm{2.5}$ & 13.8  & 13.45$\pm{1.52}$ \\ 
        Place Only(BD)       &26.5   & 25.36$\pm{1.14}$ & 14.5  & 13.60$\pm{0.40}$ \\ 
        Cap Only (BD)        & 27.8  & 26.40 $\pm{1.52}$& 18.4 & 17.60$\pm{0.65}$ \\ 
        Tag Only (BD)        &20.3   & 18.13 $\pm{2.09}$  &12.1  &12.10$\pm{0.61}$ \\  \hline
        Img+Place (BD)       &27.7  & 26.96  $\pm{0.65}$& 16.5  & 16.00$\pm{0.41}$ \\ 
        Img+Cap (BD)         &26.5   & 24.43 $\pm{1.14}$   &15.0  &14.56 $\pm{0.31}$\\  
        Img+Tag (BD)         & 31.4 & 29.96  $\pm{1.47}$ & 20.1 & 18.96$\pm{1.08}$\\\hline
        Img+Place+Cap (BD)  &28.7  &27.86   $\pm{0.74}$  & 18.1&15.56$\pm{1.77}$  \\
        Img+Place+Tag (BD)   &30.6  & 28.46  $\pm{1.58}$ &18.5  & 17.60 $\pm{0.73}$\\
        Img+Cap+Tag (BD)     &37.3  & 36.43  $\pm{1.15}$  &21.7  &  20.70$\pm{0.49}$    \\  \hline 
        MC-SMN(Img+Place+Cap+Tag(w/o Dropout))   & 33.3 & 33.33 $\pm{0.00}$ & 21.1 &21.10 $\pm{0.00}$   \\
        MC-BMN (Img+Place+Cap+Tag (Gaussian Dropout)) & 38.6 & 35.63 $\pm{2.73}$& {22.9} &21.53 $\pm{1.06}$ \\
         MC-BMN(Img+Place+Cap+Tag(BD)) (\textbf{Ours}) & \textbf{40.7} & \textbf{38.73 $\pm{1.67}$}& \text{22.6}  &{22.03} $\pm{0.80}$ \\  \hline 
        Humans\cite{mostafazadeh2016generating} & {86.0}& --&{60.8} & --\\\hline
  \end{tabular}
  \caption{\label{tab_avg_max}Comparison with \textbf{state-of-the-art} and different combination of \textbf{Cues}. The first block consists of the SOTA methods, second block depicts the models which uses only a single type of information such as Image or Place, third block has models which take one cue along with the Image information, fourth block takes two cues along with the Image information. The second last block consists of variations of our method. First is MC-SMN (Simple Moderator Network) in which there is no dropout (w/o Dropout) at inference time as explained in section~\ref{baseline_sota} and the second one uses Gaussian dropout instead of the Bernoulli dropout (BD) which we have used across all the models.}
  \vspace{-1.5em}
\end{table*}
\vspace{-0.5em}
\subsection{Dataset}
We conduct our experiments on  Visual Question Generation (VQG) dataset~\cite{mostafazadeh2016generating}, which contains human annotated questions based on images of MS-COCO dataset. 
This dataset~\cite{mostafazadeh2016generating} was developed for generating natural and engaging questions.
It contains a total of 2500 training images, 1250 validation images, and 1250 testing images. Each image in the dataset contains five natural questions and five ground truth captions. It is worth noting that the work of~\cite{jain2017creativity} also used the questions from VQA dataset~\cite{VQA} for training purpose, whereas the work by~\cite{mostafazadeh2016generating} uses only the VQG-COCO dataset. We understand that the size of this dataset is small and there are other datasets like VQA~\cite{VQA}, Visual7W~\cite{zhu2016visual7w} and Visual Genome~\cite{krishna2017visual} which have thousands of images and questions. But, VQA questions are mainly visually grounded and literal, Visual7w questions are designed to be answerable by only the image, and questions in Visual Genome focus on cognitive tasks, making them unnatural for asking a human~\cite{mostafazadeh2016generating} and hence not suited for the VQG task.

  \vspace{-0.3em}
\subsection{Comparison with different cues}\label{variants}
The first analysis is considering the various combinations of cues such as caption and place. The comparison is provided in table~\ref{tab_avg_max}. The second block of table~\ref{tab_avg_max} depicts the models which use only a single type of information such as Image or Place. We use these models as our baseline and compare other variations of our model with the best single cue. The third block takes into consideration one cue along with the Image information, and we see an improvement of around 4\% in BLEU1 and 2\% in METEOR score. The fourth block takes two cues along with the Image information and obtains an improvement of around 10\% in BLEU and 3\% in METEOR scores. The question tags performs the best among all the 3 tags. This is reasonable as question tag can guide the type of question. The second last block consists of variations of our method. the first variation corresponds to the model in which there is no dropout at inference time and the second one uses Gaussian dropout instead of the Bernoulli dropout which we have used across all the models. As we can see, the application of dropout leads to a significant increase in the BLEU score and also Bernoulli dropout works best. We also observe that our proposed method MC-BMN gets an improvement of 13\% in BLEU and 5\% in METEOR score over the single cue baselines. Tags work well in general along with other cues than caption as it provides more precise information compared to the caption, but the performance drops significantly if only the tag information is provided as there is not much information for generating sensible questions.
While comparing the various embedding, we also evaluated various ways of integrating the different cues to obtain joint embedding. 
\begin{figure*}[ht]
\centering
\includegraphics[width=0.9\linewidth]{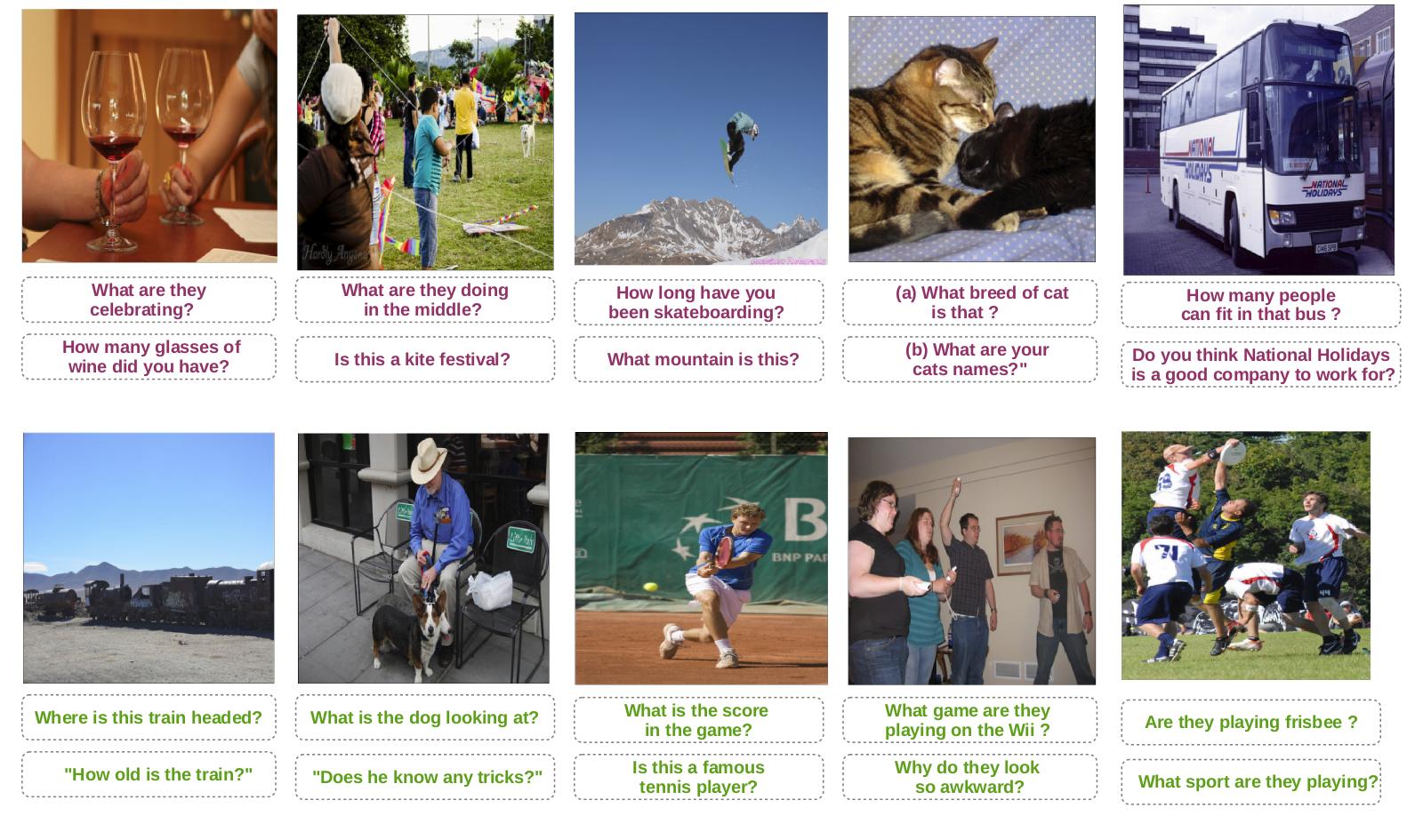} 
\vspace{-0.7em}
\caption{Examples of questions generated by our method for different images. First question in each image is generated by our method and second one is the ground truth question. More results are present in the project webpage.}
\label{fig:univerise_result}
\vspace{-1.0em}
\end{figure*}

\vspace{-0.3em}
\subsection{Comparison with state-of-the-art  methods and Ablation Analysis}\label{baseline_sota}
\vspace{-0.5em}

The comparison of our method with various state-of-the-art methods and ablation analysis is provided in table~\ref{tab_avg_max}. We observe that in terms of METEOR score, obtain an improvement of around 3\% using our proposed method over previous work by Mostafazadeh et. al~\cite{mostafazadeh2016generating} and Jain et. al~\cite{jain2017creativity}. For BLEU score the improvement is around 20\% over~\cite{mostafazadeh2016generating}, 5\% over~\cite{jain2017creativity}. But it's still quite far from human performance. 

\noindent \textbf{Ablation Analysis:} We consider different variants of our methods. These are  use of  Conventional CNN and a concatenation of the various embeddings (Multi Cue Simple Mixture (MC-SMix)), a Bayesian CNN and concatenation of the various embeddings (Multi Cue Bayesian Mixture (MC-BMix)), and the final one uses a mixture of experts along with a conventional CNN (Multi Cue Simple Moderator Network (MC-SMN)). MC-SMN actually corresponds to our MC-BMN method without dropout. Our proposed method improves upon these ablations. 
\begin{figure}[ht]
	\centering
	\includegraphics[width=\columnwidth]{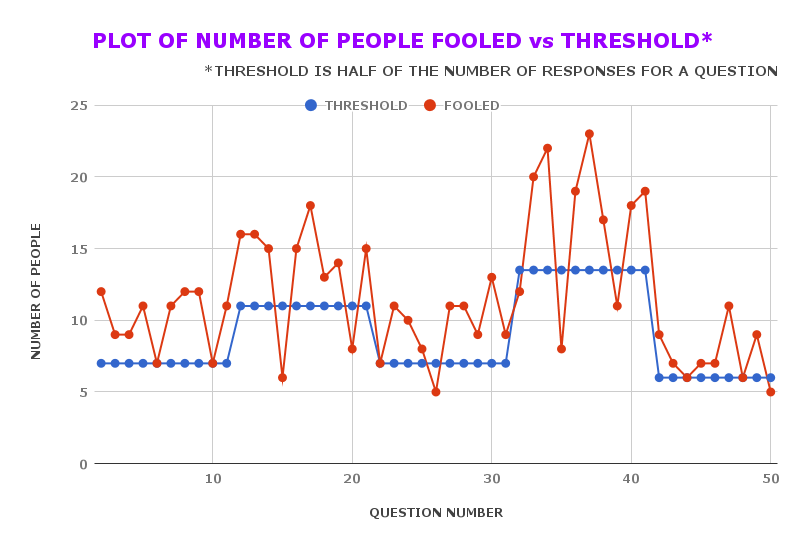}
		\vspace{-0.9em}
	\caption{Perceptual Realism Plot for human survey (section~\ref{human_survey}). The blue and red dots represent the threshold and the number of people fooled for each question respectively. Here every question has different number of responses and hence the threshold for each question is varying. Also, we are only providing the plot for 50 of 100 questions involved in the survey.}
	\label{fig:human_survey}
		\vspace{-1em}
\end{figure}
\subsection{Perceptual Realism}\label{human_survey}
A human is the best judge of the naturalness of any question; we also evaluated our proposed MC-BMN method using a `Naturalness' Turing test~\cite{Zhang_ECCV2016} on 175 people. 
People were shown an image with two questions just as in figure~\ref{fig:univerise_result} and were asked to rate the naturalness of both the questions on a scale of 1 to 5  where one means `Least Natural' and 5 is the `Most Natural.' We provided them with 100 such images from the VQG-COCO validation dataset which has 1250 images. 
Figure~\ref{fig:human_survey} indicates the number of people who were fooled (rated the generated question more or equal to the ground truth question). For the 100 images, on an average 61.8\%, people were fooled. If we provide both questions as the ground truth ones then on an average 50 \% people were fooled, and this shows that our model can generate natural questions. 


\vspace{-1.25em}
\section{Conclusion}
\vspace{-0.5em}
In this paper, we have proposed a novel solution for the problem of generating natural questions for an image. The approach relies on obtaining the advice of different Bayesian experts that are used for generating natural questions. We provide a detailed comparison with state of the art baseline methods, perform a user study to evaluate the naturalness of the questions and also ensure that the results are statistically significant. Our work introduces a principled framework to include cues for vision and language-based interaction. We aim to further validate the generalization of the approach by extending this approach to other vision and language tasks. The resulting approach has been also analysed in terms of Conventional CNN, Bayesian LSTM with product of experts and we observe that the proposed Bayesian Expert model  improved over all the other variants.
{\small
\bibliographystyle{ieee}
\bibliography{egbib}

\begin{thebibliography}{10}\itemsep=-1pt

\bibitem{VQA}
S.~Antol, A.~Agrawal, J.~Lu, M.~Mitchell, D.~Batra, C.~L. Zitnick, and
  D.~Parikh.
\newblock {VQA}: {V}isual {Q}uestion {A}nswering.
\newblock In {\em International Conference on Computer Vision (ICCV)}, 2015.

\bibitem{Aytar_TPAMI2017}
Y.~Aytar, L.~Castrejon, C.~Vondrick, H.~Pirsiavash, and A.~Torralba.
\newblock Cross-modal scene networks.
\newblock {\em IEEE transactions on pattern analysis and machine intelligence},
  2017.

\bibitem{Baldacchino_MSSP2016}
T.~Baldacchino, E.~J. Cross, K.~Worden, and J.~Rowson.
\newblock Variational bayesian mixture of experts models and sensitivity
  analysis for nonlinear dynamical systems.
\newblock {\em Mechanical Systems and Signal Processing}, 66:178--200, 2016.

\bibitem{Banerjee_ACL2005}
S.~Banerjee and A.~Lavie.
\newblock Meteor: An automatic metric for mt evaluation with improved
  correlation with human judgments.
\newblock In {\em Proc. of ACL workshop on Intrinsic and Extrinsic Evaluation
  measures for Machine Translation and/or Summarization}, volume~29, pages
  65--72, 2005.

\bibitem{Barber_NATO1998}
D.~Barber and C.~M. Bishop.
\newblock Ensemble learning in bayesian neural networks.
\newblock pages 215--238, 1998.

\bibitem{Barnard_JMLR2003}
K.~Barnard, P.~Duygulu, and D.~Forsyth.
\newblock N. de freitas, d.
\newblock {\em Blei, and MI Jordan," Matching Words and Pictures", submitted to
  JMLR}, 2003.

\bibitem{Blundell_ARX2015}
C.~Blundell, J.~Cornebise, K.~Kavukcuoglu, and D.~Wierstra.
\newblock Weight uncertainty in neural networks.
\newblock {\em arXiv preprint arXiv:1505.05424}, 2015.

\bibitem{buntine1991bayesian}
W.~L. Buntine and A.~S. Weigend.
\newblock Bayesian back-propagation.
\newblock {\em Complex systems}, 5(6):603--643, 1991.

\bibitem{visdial_eval}
P.~Chattopadhyay, D.~Yadav, V.~Prabhu, A.~Chandrasekaran, A.~Das, S.~Lee,
  D.~Batra, and D.~Parikh.
\newblock Evaluating visual conversational agents via cooperative human-ai
  games.
\newblock In {\em Proceedings of the Fifth AAAI Conference on Human Computation
  and Crowdsourcing (HCOMP)}, 2017.

\bibitem{Chen_CVPR2015}
X.~Chen and C.~Lawrence~Zitnick.
\newblock Mind's eye: A recurrent visual representation for image caption
  generation.
\newblock In {\em Proceedings of the IEEE conference on computer vision and
  pattern recognition}, pages 2422--2431, 2015.

\bibitem{visdial}
A.~Das, S.~Kottur, K.~Gupta, A.~Singh, D.~Yadav, J.~M. Moura, D.~Parikh, and
  D.~Batra.
\newblock {V}isual {D}ialog.
\newblock In {\em IEEE Conference on Computer Vision and Pattern Recognition
  (CVPR)}, 2017.

\bibitem{denker1987large}
J.~Denker, D.~Schwartz, B.~Wittner, S.~Solla, R.~Howard, L.~Jackel, and
  J.~Hopfield.
\newblock Large automatic learning, rule extraction, and generalization.
\newblock {\em Complex systems}, 1(5):877--922, 1987.

\bibitem{denker1991transforming}
J.~S. Denker and Y.~Lecun.
\newblock Transforming neural-net output levels to probability distributions.
\newblock In {\em Advances in neural information processing systems}, pages
  853--859, 1991.

\bibitem{Fang_CVPR2015}
H.~Fang, S.~Gupta, F.~Iandola, R.~Srivastava, L.~Deng, P.~Doll{\'a}r, J.~Gao,
  X.~He, M.~Mitchell, J.~Platt, et~al.
\newblock From captions to visual concepts and back.
\newblock In {\em Proceedings of the IEEE conference on computer vision and
  pattern recognition}, 2015.

\bibitem{Farhadi_ECCV2010}
A.~Farhadi, M.~Hejrati, M.~A. Sadeghi, P.~Young, C.~Rashtchian, J.~Hockenmaier,
  and D.~Forsyth.
\newblock Every picture tells a story: Generating sentences from images.
\newblock In {\em European conference on computer vision}, pages 15--29.
  Springer, 2010.

\bibitem{Fukui_arXiv2016}
A.~Fukui, D.~H. Park, D.~Yang, A.~Rohrbach, T.~Darrell, and M.~Rohrbach.
\newblock Multimodal compact bilinear pooling for visual question answering and
  visual grounding.
\newblock {\em arXiv preprint arXiv:1606.01847}, 2016.

\bibitem{Gal_ARX2015}
Y.~Gal and Z.~Ghahramani.
\newblock Bayesian convolutional neural networks with bernoulli approximate
  variational inference.
\newblock {\em arXiv preprint arXiv:1506.02158}, 2015.

\bibitem{Gal_ICML2016}
Y.~Gal and Z.~Ghahramani.
\newblock Dropout as a bayesian approximation: Representing model uncertainty
  in deep learning.
\newblock In {\em International Conference on Machine Learning (ICML)}, pages
  1050--1059, 2016.

\bibitem{Gal_NIPS2016}
Y.~Gal and Z.~Ghahramani.
\newblock A theoretically grounded application of dropout in recurrent neural
  networks.
\newblock In {\em Advances in neural information processing systems}, pages
  1019--1027, 2016.

\bibitem{GanjuCVPR17}
S.~Ganju, O.~Russakovsky, and A.~Gupta.
\newblock What's in a question: Using visual questions as a form of
  supervision.
\newblock In {\em CVPR}, 2017.

\bibitem{Graves_NIPS2011}
A.~Graves.
\newblock Practical variational inference for neural networks.
\newblock In {\em Advances in Neural Information Processing Systems (NIPS)},
  pages 2348--2356, 2011.

\bibitem{Hinton_ACM1993}
G.~E. Hinton and D.~Van~Camp.
\newblock Keeping the neural networks simple by minimizing the description
  length of the weights.
\newblock In {\em Proc. of the Conference on Computational learning theory
  (COLT)}, pages 5--13. ACM, 1993.

\bibitem{jain2017creativity}
U.~Jain, Z.~Zhang, and A.~G. Schwing.
\newblock Creativity: Generating diverse questions using variational
  autoencoders.
\newblock In {\em Proceedings of the IEEE Conference on Computer Vision and
  Pattern Recognition}, pages 6485--6494, 2017.

\bibitem{Johnson_CVPR2016}
J.~Johnson, A.~Karpathy, and L.~Fei-Fei.
\newblock Densecap: Fully convolutional localization networks for dense
  captioning.
\newblock In {\em Proceedings of the IEEE Conference on Computer Vision and
  Pattern Recognition}, pages 4565--4574, 2016.

\bibitem{Karpathy_CVPR2015}
A.~Karpathy and L.~Fei-Fei.
\newblock Deep visual-semantic alignments for generating image descriptions.
\newblock In {\em Proceedings of the IEEE conference on computer vision and
  pattern recognition}, pages 3128--3137, 2015.

\bibitem{kendall2015bayesian}
A.~Kendall, V.~Badrinarayanan, and R.~Cipolla.
\newblock Bayesian segnet: Model uncertainty in deep convolutional
  encoder-decoder architectures for scene understanding.
\newblock {\em arXiv preprint arXiv:1511.02680}, 2015.

\bibitem{Kendall_CVPR2018}
A.~Kendall, Y.~Gal, and R.~Cipolla.
\newblock Multi-task learning using uncertainty to weigh losses for scene
  geometry and semantics.
\newblock 2018.

\bibitem{Kim_ICLR2017}
J.-H. Kim, K.~W. On, W.~Lim, J.~Kim, J.-W. Ha, and B.-T. Zhang.
\newblock {Hadamard Product for Low-rank Bilinear Pooling}.
\newblock In {\em The 5th International Conference on Learning
  Representations}, 2017.

\bibitem{krishna2017visual}
R.~Krishna, Y.~Zhu, O.~Groth, J.~Johnson, K.~Hata, J.~Kravitz, S.~Chen,
  Y.~Kalantidis, L.-J. Li, D.~A. Shamma, et~al.
\newblock Visual genome: Connecting language and vision using crowdsourced
  dense image annotations.
\newblock {\em International Journal of Computer Vision}, 123(1):32--73, 2017.

\bibitem{Kulkarni_CVPR2011}
G.~Kulkarni, V.~Premraj, S.~Dhar, S.~Li, Y.~Choi, A.~C. Berg, and T.~L. Berg.
\newblock Baby talk: Understanding and generating image descriptions.
\newblock In {\em Proceedings of the 24th CVPR}. Citeseer, 2011.

\bibitem{kurmi_cvpr2019attending}
V.~K. Kurmi, S.~Kumar, and V.~P. Namboodiri.
\newblock Attending to discriminative certainty for domain adaptation.
\newblock In {\em Proceedings of the IEEE Conference on Computer Vision and
  Pattern Recognition}, pages 491--500, 2019.

\bibitem{Lin_ACL2004}
C.-Y. Lin.
\newblock Rouge: A package for automatic evaluation of summaries.
\newblock In {\em Text summarization branches out:Proceedings of the ACL-04
  workshop}, 2004.

\bibitem{Lin_ECCV2014}
T.-Y. Lin, M.~Maire, S.~Belongie, J.~Hays, P.~Perona, D.~Ramanan,
  P.~Doll{\'a}r, and C.~L. Zitnick.
\newblock Microsoft coco: Common objects in context.
\newblock In {\em European Conference on Computer Vision}, pages 740--755.
  Springer, 2014.

\bibitem{Ma_AAAI2016}
L.~Ma, Z.~Lu, and H.~Li.
\newblock Learning to answer questions from image using convolutional neural
  network.
\newblock In {\em Thirtieth AAAI Conference on Artificial Intelligence}, 2016.

\bibitem{mackay1992bayesian}
D.~J. MacKay.
\newblock Bayesian interpolation.
\newblock {\em Neural computation}, 4(3):415--447, 1992.

\bibitem{malinin2018predictive}
A.~Malinin and M.~Gales.
\newblock Predictive uncertainty estimation via prior networks.
\newblock In {\em Advances in Neural Information Processing Systems}, pages
  7047--7058, 2018.

\bibitem{Malinowski_NIPS2014}
M.~Malinowski and M.~Fritz.
\newblock A multi-world approach to question answering about real-world scenes
  based on uncertain input.
\newblock In {\em Advances in Neural Information Processing Systems (NIPS)},
  2014.

\bibitem{mostafazadeh2016generating}
N.~Mostafazadeh, I.~Misra, J.~Devlin, M.~Mitchell, X.~He, and L.~Vanderwende.
\newblock Generating natural questions about an image.
\newblock In {\em Proceedings of the 54th Annual Meeting of the Association for
  Computational Linguistics (Volume 1: Long Papers)}, volume~1, pages
  1802--1813, 2016.

\bibitem{neal1993bayesian}
R.~M. Neal.
\newblock Bayesian learning via stochastic dynamics.
\newblock In {\em Advances in neural information processing systems}, pages
  475--482, 1993.

\bibitem{neal2012bayesian}
R.~M. Neal.
\newblock {\em Bayesian learning for neural networks}, volume 118.
\newblock Springer Science \& Business Media, 2012.

\bibitem{Noh_CVPR2016}
H.~Noh, P.~Hongsuck~Seo, and B.~Han.
\newblock Image question answering using convolutional neural network with
  dynamic parameter prediction.
\newblock In {\em Proceedings of the IEEE Conference on Computer Vision and
  Pattern Recognition}, pages 30--38, 2016.

\bibitem{Papineni_ACL2002}
K.~Papineni, S.~Roukos, T.~Ward, and W.-J. Zhu.
\newblock Bleu: a method for automatic evaluation of machine translation.
\newblock In {\em Proceedings of the 40th annual meeting on association for
  computational linguistics}, pages 311--318. Association for Computational
  Linguistics, 2002.

\bibitem{patro2018multimodal}
B.~N. Patro, S.~Kumar, V.~K. Kurmi, and V.~Namboodiri.
\newblock Multimodal differential network for visual question generation.
\newblock In {\em Proceedings of the 2018 Conference on Empirical Methods in
  Natural Language Processing}, pages 4002--4012, 2018.

\bibitem{Patro_2019_ICCV}
B.~N. Patro, M.~Lunayach, S.~Patel, and V.~P. Namboodiri.
\newblock U-cam: Visual explanation using uncertainty based class activation
  maps.
\newblock In {\em The IEEE International Conference on Computer Vision (ICCV)},
  October 2019.

\bibitem{Ren_NIPS2015}
M.~Ren, R.~Kiros, and R.~Zemel.
\newblock Exploring models and data for image question answering.
\newblock In {\em Advances in Neural Information Processing Systems (NIPS)},
  pages 2953--2961, 2015.

\bibitem{selvaraju2017grad}
R.~R. Selvaraju, M.~Cogswell, A.~Das, R.~Vedantam, D.~Parikh, and D.~Batra.
\newblock Grad-cam: Visual explanations from deep networks via gradient-based
  localization.
\newblock In {\em Computer Vision (ICCV), 2017 IEEE International Conference
  on}, pages 618--626. IEEE, 2017.

\bibitem{simonyan_arxiv2014}
K.~Simonyan and A.~Zisserman.
\newblock Very deep convolutional networks for large-scale image recognition.
\newblock {\em arXiv preprint arXiv:1409.1556}, 2014.

\bibitem{Socher_TACL2014}
R.~Socher, A.~Karpathy, Q.~V. Le, C.~D. Manning, and A.~Y. Ng.
\newblock Grounded compositional semantics for finding and describing images
  with sentences.
\newblock {\em Transactions of the Association of Computational Linguistics},
  2(1):207--218, 2014.

\bibitem{srivastava2014dropout}
N.~Srivastava, G.~Hinton, A.~Krizhevsky, I.~Sutskever, and R.~Salakhutdinov.
\newblock Dropout: a simple way to prevent neural networks from overfitting.
\newblock {\em The Journal of Machine Learning Research}, 15(1):1929--1958,
  2014.

\bibitem{Sutskever_NIPS2014}
I.~Sutskever, O.~Vinyals, and Q.~V. Le.
\newblock Sequence to sequence learning with neural networks.
\newblock In {\em Advances in Neural Information Processing Systems (NIPS)},
  pages 3104--3112, 2014.

\bibitem{tishby1989consistent}
N.~Tishby, E.~Levin, and S.~A. Solla.
\newblock Consistent inference of probabilities in layered networks:
  Predictions and generalization.
\newblock In {\em IJCNN International Joint Conference on Neural Networks},
  volume~2, pages 403--409. IEEE New York, 1989.

\bibitem{Ueda_IEEE2000}
N.~Ueda and Z.~Ghahramani.
\newblock Optimal model inference for bayesian mixture of experts.
\newblock In {\em IEEE Workshop on Neural Networks for Signal Processing X},
  volume~1, pages 145--154. IEEE, 2000.

\bibitem{Vedantam_CVPR2015}
R.~Vedantam, L.~Zitnick, and D.~Parikh.
\newblock Cider: Consensus-based image description evaluation.
\newblock In {\em IEEE Conference on Computer Vision and Pattern Recognition
  (CVPR)}, pages 4566--4575, 2015.

\bibitem{Velivckovic_SSCI2016}
P.~Veli{\v{c}}kovi{\'c}, D.~Wang, N.~D. Lane, and P.~Li{\`o}.
\newblock X-cnn: Cross-modal convolutional neural networks for sparse datasets.
\newblock In {\em Computational Intelligence (SSCI), 2016 IEEE Symposium Series
  on}, pages 1--8. IEEE, 2016.

\bibitem{Vijayakumar_2016diverse}
A.~K. Vijayakumar, M.~Cogswell, R.~R. Selvaraju, Q.~Sun, S.~Lee, D.~Crandall,
  and D.~Batra.
\newblock Diverse beam search: Decoding diverse solutions from neural sequence
  models.
\newblock {\em arXiv preprint arXiv:1610.02424}, 2016.

\bibitem{Vinyals_CVPR2015}
O.~Vinyals, A.~Toshev, S.~Bengio, and D.~Erhan.
\newblock Show and tell: A neural image caption generator.
\newblock In {\em Proceedings of the IEEE Conference on Computer Vision and
  Pattern Recognition}, pages 3156--3164, 2015.

\bibitem{Xu_ICML2015}
K.~Xu, J.~Ba, R.~Kiros, K.~Cho, A.~Courville, R.~Salakhudinov, R.~Zemel, and
  Y.~Bengio.
\newblock Show, attend and tell: Neural image caption generation with visual
  attention.
\newblock In {\em International Conference on Machine Learning}, pages
  2048--2057, 2015.

\bibitem{Yan_ECCV2016}
X.~Yan, J.~Yang, K.~Sohn, and H.~Lee.
\newblock Attribute2image: Conditional image generation from visual attributes.
\newblock In {\em European Conference on Computer Vision}, pages 776--791.
  Springer, 2016.

\bibitem{Yang_arXiv2015}
Y.~Yang, Y.~Li, C.~Fermuller, and Y.~Aloimonos.
\newblock Neural self talk: Image understanding via continuous questioning and
  answering.
\newblock {\em arXiv preprint arXiv:1512.03460}, 2015.

\bibitem{Yang_CVPR2016}
Z.~Yang, X.~He, J.~Gao, L.~Deng, and A.~Smola.
\newblock Stacked attention networks for image question answering.
\newblock In {\em Proceedings of the IEEE Conference on Computer Vision and
  Pattern Recognition}, pages 21--29, 2016.

\bibitem{Yu_ICCV2015}
L.~Yu, E.~Park, A.~C. Berg, and T.~L. Berg.
\newblock Visual madlibs: Fill in the blank description generation and question
  answering.
\newblock In {\em Computer Vision (ICCV), 2015 IEEE International Conference
  on}, pages 2461--2469. IEEE, 2015.

\bibitem{Yuksel_IEEETransNNLS2012}
S.~E. Yuksel, J.~N. Wilson, and P.~D. Gader.
\newblock Twenty years of mixture of experts.
\newblock {\em IEEE Transactions on Neural Networks and Learning Systems},
  23(8):1177--1193, 2012.

\bibitem{Zhang_ECCV2016}
R.~Zhang, P.~Isola, and A.~A. Efros.
\newblock Colorful image colorization.
\newblock In {\em European Conference on Computer Vision}, pages 649--666.
  Springer, 2016.

\bibitem{Zhou_PAMI2017}
B.~Zhou, A.~Lapedriza, A.~Khosla, A.~Oliva, and A.~Torralba.
\newblock Places: A 10 million image database for scene recognition.
\newblock {\em IEEE Transactions on Pattern Analysis and Machine Intelligence},
  2017.

\bibitem{Zhou_arXiv2015}
B.~Zhou, Y.~Tian, S.~Sukhbaatar, A.~Szlam, and R.~Fergus.
\newblock Simple baseline for visual question answering.
\newblock {\em arXiv preprint arXiv:1512.02167}, 2015.

\bibitem{zhu2016visual7w}
Y.~Zhu, O.~Groth, M.~Bernstein, and L.~Fei-Fei.
\newblock Visual7w: Grounded question answering in images.
\newblock In {\em Proceedings of the IEEE Conference on Computer Vision and
  Pattern Recognition}, pages 4995--5004, 2016.

\end{thebibliography}
}

\end{document}